# Axiomatizing Causal Reasoning*

Joseph Y. Halpern
Cornell University
Computer Science Department
Ithaca, NY 14853
halpern@.cornell.edu
http://www.cs.cornell.edu/home/halpern

## Abstract

Causal models defined in terms of a collection of equations, as defined by Pearl, are axiomatized here. Axiomatizations are provided for three successively more general classes of causal models: (1) the class of recursive theories (those without feedback), (2) the class of theories where the solutions to the equations are unique, (3) arbitrary theories (where the equations may not have solutions and, if they do, they are not necessarily unique). It is shown that to reason about causality in the most general third class, we must extend the language used by Galles and Pearl. In addition, the complexity of the decision procedures is examined for all the languages and classes of models considered.

## 1 INTRODUCTION

The important role of causal reasoning—in prediction, explanation, and counterfactual reasoning—has been argued eloquently in a number of recent papers and books [Chajewska and Halpern 1997; Heckerman and Shachter 1995; Henrion and Druzdzel 1990; Druzdzel and Simon 1993; Pearl 1995; Pearl and Verma 1991; Spirtes, Glymour, and Scheines 1993]. If we are to reason about causality, then it is certainly useful to find axioms that characterize such reasoning. The way we go about axiomatizing causal reasoning depends on two critical factors:

- how we model causality, and
- the language that we use to reason about it.

In this paper, I consider one approach to modeling causality, using *structural equations*. The use of structural equations as a model for causality is standard in the social sciences, and seems to go back to the work of Sewall Wright in the 1920s (see [Goldberger 1972] for a discussion); the particular framework that I use here is due to Pearl [1995]. Galles and Pearl [1997] introduce some axioms for causal reasoning in this framework; in [Galles and Pearl 1998], they provide a complete axiomatic characterization of reasoning about causality in this framework, under the strong assumption that there is a fixed, given *causal ordering* of the equations. Roughly speaking, this means there is a way of ordering the variables that appear in the equations and we have explicit axioms that say $X_j$ has no influence of $X_i$ if $X_j > X_i$ in this causal ordering.

In this paper, I extend the results of Galles and Pearl by providing a complete axiomatic characterization for three increasingly general classes of causal models (defined by structural equations):

1. the class of recursive theories (those without feedback—this generalizes the situation considered by Galles and Pearl [1998], since every fixed causal ordering of the variables gives rise to a recursive theory),

2. the class of theories where the solutions to the equations are unique,

3. arbitrary theories (where the equations may not have solutions and, if they do, they are not necessarily unique).

In the process, I clarify some problems in the Galles-Pearl completeness proof that arise from the lack of propositional connectives (particularly disjunction) in the language they consider and, more generally, highlight the role of the language in reasoning about causality. I also consider the complexity of the decision problem for all these languages and classes of models.

The rest of the paper is organized as follows. In Section 2, I give the syntax and semantics of the languages I will be considering and review the definition of modifiable causal models. In Section 3, I present the complete axiomatizations. In Section 4 I consider the complexity of the decision procedure. I conclude in Section 5. Some proofs are given in the appendix.

*This work was supported in part by NSF under grant -96-25901 and by the Air Force Office of Scientific Research under grant F49620-96-1-0323.

## 2  SYNTAX AND SEMANTICS

An axiomatization is given with respect to a particular language and a class of models, so we need to make both precise. Both the language and models I use are based on those considered by Galles and Pearl [1997, 1998]. To make comparisons easier, I use their notation as much as possible. I start with the semantic model, since it motivates some of the choices in the syntax, then give the syntax, and finally define the semantics of formulas.

**Causal Models**  The basic picture here is that we are interested in the values of random variables, some of which have a causal effect on others. This effect is modeled by a set of *structural equations*. In practice, it seems useful to split the random variables into two sets, the *exogenous* variables, whose values are determined by factors outside the model, and the *endogenous* variables. It is these endogenous variables whose values are described by the structural equations.

More formally, a *signature* $\mathcal{S}$ is a tuple $(\mathcal{U}, \mathcal{X}, \{V_Y : Y \in \mathcal{U} \cup \mathcal{X}\})$, where $\mathcal{U}$ is a finite set of exogenous variables, $\mathcal{X}$ is a finite set of endogenous variables, and $V_Y$ is a set of possible values for each random variable $Y \in \mathcal{U} \cup \mathcal{X}$. Unless explicitly noted otherwise, I assume that $V_Y$ is a *finite* set for each $Y \in \mathcal{U} \cup \mathcal{X}$ and $|V_X| \geq 2$. The assumption that $\mathcal{X}$ is finite is relatively innocuous; as we shall see, the assumption that $V_Y$ is finite has more of an impact on both the axioms and decision procedures. The assumption that $|V_X| \geq 2$ allows us to ignore the trivial situation where $|V_X| = 1$. If $|V_X| = 1$, we can just remove the variable $X$ from the signature without loss of expressiveness.

A *causal model* over signature $\mathcal{S}$ is a tuple $T = (\mathcal{S}, \{F_X : X \in \mathcal{X}\})$, where $\{F_X : X \in \mathcal{X}\}$ is a set of *(modifiable) structural equations*. The structural equation $F_X$ is a function mapping $(\times_{U \in \mathcal{U}} V_U) \times (\times_{Y \in \mathcal{X} - \{X\}} V_Y) \to V_X$. $F_X$ tells us the value of $X$ given the values of all the other variables in $\mathcal{U} \cup \mathcal{X}$. Because $F_X$ is a function, there is a unique value of $X$ once we have set all the other variables. Notice we have such functions only for the endogenous variables. The exogenous variables are taken as given; it is their effect on the endogenous variables (and the effect of the endogenous variables on each other) that we are modeling with the structural equations.

Given a causal model $T$ over signature $\mathcal{S}$, a (possibly empty) set $\vec{X} \subseteq \mathcal{X}$, and values $\vec{x}$ for the variables in $\vec{X}$, we can define a new causal model denoted $T_{\vec{X} \leftarrow \vec{x}}(\vec{u})$ over the signature $\mathcal{S}' = (\emptyset, \mathcal{X} - \vec{X}, \{V_Y : Y \in \mathcal{X} - \vec{X}\})$. Intuitively, this is the causal model that results when the variables in $\vec{X}$ are set to $\vec{x}$ and the variables in $\mathcal{U}$ are set to $\vec{u}$. Formally, $T_{\vec{X} \leftarrow \vec{x}}(\vec{u}) = (\mathcal{S}', \{F_Y^{\vec{x},\vec{u}} : X \in \mathcal{X} - \vec{X}\})$, where $F_Y^{\vec{x},\vec{u}}$ is obtained from $F_Y$ by setting the values of the variables in $\vec{X}$ to $\vec{x}$ and the values of the variables in $\mathcal{U}$ to $\vec{u}$.

Notice that, in general, there may not be a unique vector of values that simultaneously satisfies the equations in $T_{\vec{X} \leftarrow \vec{x}}(\vec{u})$; indeed, there may not be a solution at all. One special case where there is guaranteed to be such a unique solution is if there is some total ordering $\prec$ of the variables in $\mathcal{X}$ such that if $X \prec Y$, then $F_X$ is independent of the values of $Y$; i.e., $F_X(\ldots, y, \ldots) = F_X(\ldots, y', \ldots)$ for all $y, y' \in V_Y$. In this case, the causal model is said to be *recursive* or *acyclic*. Intuitively, if the theory is recursive, there is no feedback. If $X \prec Y$, then the value of $X$ may affect the value of $Y$, but the value of $Y$ has no effect on the value of $X$.

It should be clear that if $T$ is a recursive theory, then there is always a unique solution to the equations in $T_{\vec{X} \leftarrow \vec{x}}(\vec{u})$, for all $\vec{X}$, $\vec{x}$, and $\vec{u}$. (We simply solve for the variables in the order given by $\prec$.) On the other hand, as the following example shows, it is not hard to construct nonrecursive theories for which there is always a unique solution to the equations that arise.

**Example 2.1:** Let $\mathcal{S} = (\emptyset, \{X, Y\}, \{V_X, V_Y\})$, where $V_X = V_Y = \{-1, 0, 1\}$, and let $T = (\mathcal{S}, \{F_X, F_Y\})$, where $F_X$ is characterized by the equation $X = Y$ and $F_Y$ is characterized by the equation $Y = -X$. Clearly $T$ is not recursive; the value of $X$ depends on the value of $Y$ and the value of $Y$ depends on that of $X$. Nevertheless, it is easy to see that $T$ has the unique solution $X = 0, Y = 0$, $T_{X \leftarrow x}$ has the unique solution $Y = -x$, and $T_{Y \leftarrow y}$ has the unique solution $X = y$. ∎

In this paper, I consider three successively more general classes of causal models for a given signature $\mathcal{S} = (\mathcal{U}, \mathcal{X}, \{V_Y : Y \in \mathcal{X} \cup \mathcal{Y}\})$.

1. $\mathcal{T}_{\text{rec}}(\mathcal{S})$: the class of recursive causal models over signature $\mathcal{S}$

2. $\mathcal{T}_{\text{uniq}}(\mathcal{S})$: the class of causal models $T$ over $\mathcal{S}$ where for all $\vec{X} \subseteq \mathcal{X}$, $\vec{x}$, and $\vec{u}$, the equations in $T_{\vec{X} \leftarrow \vec{x}}(\vec{u})$ have a unique solution,

3. $\mathcal{T}(\mathcal{S})$: the class of all causal models over $\mathcal{S}$.

I often omit the signature $\mathcal{S}$ when it is clear from context or irrelevant, but the reader should bear in mind its important role.

Why should we be interested in causal models that do not possess unique solutions? Are there real causal systems that do not possess unique solutions? The issue of whether nonrecursive system can be given a causal interpretation is discussed at some length by Strotz and Wold [1960]. They argue that there are reasonable ways of interpreting causal interpretations where the answer is yes. It is not hard to see that under these interpretations, there may well be more than one solution to the equations. Perhaps the best way to view such equations is to think of the variables in $\mathcal{X}$ as being mutually interdependent; changing any one of them may cause a change in the others. (Think of demand and supply in economics or populations of rabbits and wolves.) The solutions to the equations then represent equilibrium situations. If there is more than one equilibrium, there will be more than one solution to the equations.

**Syntax:** I focus here on two languages. Both languages are parameterized by a signature $\mathcal{S}$. The first language, $\mathcal{L}^+(\mathcal{S})$, borrows ideas from dynamic logic [Harel 1979]. Again, I often write $\mathcal{L}^+$ rather than $\mathcal{L}^+(\mathcal{S})$ (and similarly for the other languages defined below) to simplify the notation. A *basic causal formula* is one of the form $[Y_1 \leftarrow y_1, \ldots, Y_k \leftarrow y_k]\varphi$, where $\varphi$ is a Boolean combination of formulas of the form $X(\vec{u}) = x$, $Y_1, \ldots, Y_k, X$ are variables in $\mathcal{X}$, $Y_1, \ldots, Y_k$ are distinct, $x \in V_X$, and $\vec{u}$ is a vector of values for all the variables in $\mathcal{U}$. I typically abbreviate such a formula as $[\vec{Y} \leftarrow \vec{y}]\varphi$. The special case where $k = 0$ (which is allowed) is abbreviated as $[true]\varphi$. $[\vec{Y} \leftarrow \vec{y}]\varphi$ can be interpreted as "in all possible solutions to the structural equations obtained after setting $Y_i$ to $y_i$, $i = 1, \ldots, k$, if the exogenous variables are set to $\vec{u}$, then the value of random variable $X$ will be $x$". As we shall see, this formula is true in a causal model $T$ if in all solutions to the equations in $T_{\vec{Y} \leftarrow \vec{y}}((u))$, the random variable $X$ has value $x$. Note that this formula is trivially true if there are no solutions to the structural equations. A *causal formula* is a Boolean combination of basic causal formulas.

Just as with dynamic logic, we can also define the formula $\langle \vec{Y} \leftarrow \vec{y} \rangle(X(\vec{u}) = x)$ to be an abbreviation of $\neg[\vec{Y} \leftarrow \vec{y}]\neg(X(\vec{u}) = x)$. $\langle \vec{Y} \leftarrow \vec{y} \rangle(X(\vec{u}) = x)$ is the dual of $[\vec{Y} \leftarrow \vec{y}](X(\vec{u}) = x)$. Thus, for example, $\langle \vec{Y} \leftarrow \vec{y} \rangle true$ is true if there is some solution to the equations obtained by setting $Y_i$ to $y_i$, $i = 1, \ldots, k$ (since $[\vec{Y} \leftarrow \vec{y}]false$ says that in every solution to the equations obtained by setting $Y_i$ to $y_i$, *false* is true, and thus holds exactly if the equations have no solution).

Let $\mathcal{L}_{\text{uniq}}(\mathcal{S})$ be the sublanguage of $\mathcal{L}^+(\mathcal{S})$ which consists of Boolean combinations of formulas of the form $[\vec{Y} \leftarrow \vec{y}]X(\vec{u}) = x$. Thus, the difference between $\mathcal{L}_{\text{uniq}}$ and $\mathcal{L}^+$ is that in $\mathcal{L}_{\text{uniq}}$, only $X(\vec{u}) = x$ is allowed after $[\vec{Y} \leftarrow \vec{y}]$, while in $\mathcal{L}^+$, arbitrary Boolean combinations of formulas of the form $X(\vec{u}) = x$ are allowed. As we shall see, for reasoning about causality in $\mathcal{T}_{\text{uniq}}$, the language $\mathcal{L}_{\text{uniq}}$ is adequate, since it is equivalent in expressive power to $\mathcal{L}^+$. However, this is no longer the case when reasoning about causality in $\mathcal{T}$.

Following Galles and Pearl's notation, I often write $[\vec{Y} \leftarrow \vec{y}]X(\vec{u}) = x$ as $X_{Y_1 \leftarrow y_1, \ldots, Y_k \leftarrow y_k}(\vec{u}) = x$. If the variables $Y_1, \ldots, Y_k$ are clear from context or irrelevant, I further abbreviate this as $X_{\vec{y}}(\vec{u}) = x$. (This is actually the notation used by Galles and Pearl.) If $k = 0$ (i.e., $\vec{Y}$ is the empty sequence), following Galles and Pearl, I write $X_\emptyset(\vec{u}) = x$. Let $\mathcal{L}_{\text{GP}}(\mathcal{S})$ be the sublanguage of $\mathcal{L}_{\text{uniq}}(\mathcal{S})$ consisting of just conjunctions of formulas of the form $X_{\vec{y}}(\vec{u}) = x$. In particular, it does not contain disjunctions of negations of such formulas. Although Galles and Pearl [1998] are not explicit about the language they are using, it seems to be $\mathcal{L}_{\text{GP}}$.[1]

**Semantics** A formula in $\mathcal{L}^+(\mathcal{S})$ is true or false in a causal model in $\mathcal{T}(\mathcal{S})$. As usual, we write $T \models \varphi$ if the causal formula $\varphi$ is true in causal model $T$. For a basic causal formula, we have $T \models [\vec{Y} \leftarrow \vec{y}](X(\vec{u}) = x)$ if in all solutions to $T_{\vec{Y} \leftarrow \vec{y}}(\vec{u})$ (i.e., in all vectors of values for the variables in $\mathcal{X} - \vec{X}$ that simultaneously satisfy all the equations $F'_Y$, for $Y \in \mathcal{X} - \vec{X}$), the variable $X$ has value $x$. We define the truth value of arbitrary causal formulas, which are just Boolean combinations of basic causal formulas, in the obvious way:

- $T \models \varphi_1 \wedge \varphi_2$ if $T \models \varphi_1$ and $T \models \varphi_2$
- $T \models \neg \varphi$ if $T \not\models \varphi$.

As usual, we say that a formula $\varphi$ is *valid* with respect to a class $\mathcal{T}'$ of causal models if $T \models \varphi$ for all $T \in \mathcal{T}'$.

I can now make precise the earlier claim that in $\mathcal{T}_{\text{uniq}}$ (and hence $\mathcal{T}_{\text{rec}}$), the language $\mathcal{L}_{\text{uniq}}$ is just as expressive as the full language $\mathcal{L}^+$.

**Lemma 2.2:** *The following formulas are valid in $\mathcal{T}_{\text{uniq}}$:*

*(a)* $\mathcal{T}_{\text{uniq}} \models [\vec{Y} \leftarrow \vec{y}](\varphi \vee \psi) \Leftrightarrow [\vec{Y} \leftarrow \vec{y}]\varphi \vee [\vec{Y} \leftarrow \vec{y}]\psi,$

*(b)* $\mathcal{T}_{\text{uniq}} \models [\vec{Y} \leftarrow \vec{y}](\varphi \wedge \psi) \Leftrightarrow [\vec{Y} \leftarrow \vec{y}]\varphi \wedge [\vec{Y} \leftarrow \vec{y}]\psi,$

*(c)* $\mathcal{T}_{\text{uniq}} \models [\vec{Y} \leftarrow \vec{y}]\neg \varphi \Leftrightarrow \neg[\vec{Y} \leftarrow \vec{y}]\varphi.$

*Hence, in $\mathcal{T}_{\text{uniq}}$, every formula in $\mathcal{L}^+$ is equivalent to a formula in $\mathcal{L}_{\text{uniq}}$.*

**Proof:** Straightforward; left to the reader. ■

Note that it follows from these equivalences that in $\mathcal{T}_{\text{uniq}}$, $[\vec{Y} \leftarrow \vec{y}]\varphi$ is equivalent to $\langle \vec{Y} \leftarrow \vec{y} \rangle \varphi$. It is also worth noting that Lemma 2.2(b) holds in arbitrary causal models in $\mathcal{T}$, not just in $\mathcal{T}_{\text{uniq}}$. However, parts (a) and (c) do not, as the following example shows.

**Example 2.3:** Let $\mathcal{S} = (\emptyset, \{X, Y\}, \{V_X, V_Y\})$, where $V_X = V_Y = \{0, 1\}$; let $T = (\mathcal{S}, \{F_X, F_Y\})$, where $F_X$ is characterized by the equation $X = Y$ and $F_Y$ is characterized by the equation $Y = X$. Clearly $T \notin \mathcal{T}_{\text{uniq}}$; both $(0,0)$ and $(1,1)$ are solutions to $T$. It is easy to see that $T \models [true](X = 0 \vee X = 1) \wedge \neg[true](X = 0) \wedge \neg[true](X = 1)$ and $T \models \neg[true]\neg(X = 1) \wedge \neg[true]\neg(X = 1)$, showing that neither part (a) nor part (c) of Lemma 2.2 hold in $T$. ■

---

[1] This was confirmed by Judea Pearl [private communication, 1997].

## 3  COMPLETE AXIOMATIZATIONS

I briefly recall some standard definitions from logic. An *axiom system* AX consists of a collection of *axioms* and *inference rules*. An axiom is a formula (in some predetermined language $\mathcal{L}$), and an inference rule has the form "from $\varphi_1, \ldots, \varphi_k$ infer $\psi$," where $\varphi_1, \ldots, \varphi_k, \psi$ are formulas in $\mathcal{L}$. A *proof* in AX consists of a sequence of formulas in $\mathcal{L}$, each of which is either an axiom in AX or follows by an application of an inference rule. A proof is said to be a *proof of the formula* $\varphi$ if the last formula in the proof is $\varphi$. We say $\varphi$ is *provable in AX*, and write AX $\vdash \varphi$, if there is a proof of $\varphi$ in AX; similarly, we say that $\varphi$ is *consistent with AX* if $\neg\varphi$ is not provable in AX.

An axiom system AX is said to be *sound* for a language $\mathcal{L}$ with respect to a class $\mathcal{T}'$ of causal models if every formula in $\mathcal{L}$ provable in AX is valid with respect to $\mathcal{T}'$. The system AX is *complete* for $\mathcal{L}$ with respect to $\mathcal{T}'$ if every formula in $\mathcal{L}$ that is valid with respect to $\mathcal{T}'$ is provable in AX.

We now want to find axioms that characterize the classes of causal models in which we are interested, namely $\mathcal{T}_{\text{rec}}$, $\mathcal{T}_{\text{uniq}}$, and $\mathcal{T}$. To deal with $\mathcal{T}_{\text{rec}}$, it is helpful to define $Y \rightsquigarrow Z$, read "$Y$ affects $Z$", as an abbreviation for the formula $\vee_{\vec{X} \subset \mathcal{X}, \vec{x} \in \Pi_{X \in \mathcal{X}} V_X, y \in V_y, \vec{u} \in \Pi_{U \in \mathcal{U}} V_U, z \neq z' \in V_Z} (Z_{\vec{x}y}(\vec{u}) = z' \wedge Z_{\vec{x}}(\vec{u}) = z)$. Thus, $Y$ affects $Z$ if there is some setting of the exogenous variables and some other endogenous variables for which changing the value of $Y$ changes the value of $Z$. This definition is used in axiom C6 below, which characterizes recursiveness.

Consider the following axioms:

C0. All instances of propositional tautologies.

C1. $X_{\vec{y}}(\vec{u}) = x \Rightarrow X_{\vec{y}}(\vec{u}) \neq x'$ if $x, x' \in V_X, x \neq x'$
(equality)

C2. $\vee_{x \in V_X} X_{\vec{y}}(\vec{u}) = x$    (definiteness)

C3. $(W_{\vec{x}}(\vec{u}) = w \wedge Y_{\vec{x}}(\vec{u}) = y) \Rightarrow Y_{\vec{x}w}(\vec{u}) = y$
(composition)

C4. $X_{x\vec{w}}(\vec{u}) = x$    (effectiveness)

C5. $Y_{\vec{x}w}(\vec{u}) = y \wedge W_{\vec{x}y}(\vec{u}) = w \Rightarrow Y_{\vec{x}}(\vec{u}) = y$
(reversibility)

C6. $(X_0 \rightsquigarrow X_1 \wedge \ldots \wedge X_{k_1} \rightsquigarrow X_k) \Rightarrow \neg(X_k \rightsquigarrow X_0)$
(recursiveness)

We have one rule of inference:

MP. From $\varphi$ and $\varphi \Rightarrow \psi$, infer $\psi$    (modus ponens)

C1 just states an obvious property of equality: if $X = x$ for every solution of the equations in $T_{\vec{x}}(\vec{u})$, then we cannot have $X = x'$, if $x' \neq x$.[2] In a richer language, this could have been expressed as $(X_{\vec{y}}(\vec{u}) = x \wedge X_{\vec{y}}(\vec{u}) = x') \Rightarrow (x = x')$, but this formula is not in $\mathcal{L}^+$ (since $\mathcal{L}^+$ does not include expressions such as $x' = x$). C2 states that there is some value $x \in V_X$ which is the value of $X$ in all solutions to the equations in $T_{\vec{x}}(\vec{u})$. C2 is not valid in $\mathcal{T}$, but it is valid in $\mathcal{T}_{\text{uniq}}$. Note that in stating C2, I am making use of the fact that $V_X$ is finite (otherwise C2 would involve an infinite disjunction, and would no longer be a formula in $\mathcal{L}_{\text{uniq}}$). In fact, as I show in the full paper, if we allow signatures where the sets $V_X$ are infinite, we only include C2 for those random variables $X$ such that $V_X$ is finite.[3] C3–C5 were introduced by Galles and Pearl [1997, 1998], as were their names. Roughly speaking, C3 says that if the value of $W$ is $w$ in all solutions to the equations $T_{\vec{x}}(\vec{u})$, then all solutions to the equations in $T_{\vec{x}w}(\vec{u})$ are the same as the solutions to the equations in $T_{\vec{x}}(\vec{u})$. C3 is valid in $\mathcal{T}$ as well as $\mathcal{T}_{\text{uniq}}$. As we shall see, a variant of C3 (obtained by replacing "all" by "some") is also valid in $\mathcal{T}$. C4 simply says that in all solutions obtained after setting $X$ to $x$, the value of $X$ is $x$. C5 is perhaps the least obvious of these axioms; the proof of its soundness is not at all straightforward. It says that if setting $\vec{X}$ to $\vec{x}$ and $W$ to $w$ results in $Y$ having value $y$ and setting $\vec{X}$ to $\vec{x}$ and $Y$ to $y$ results in $W$ having value $w$, then $Y$ must already have value when we set $\vec{X}$ to $x$ (and $W$ must already have value $w$).

Finally, it is easy to see that C6 holds in recursive models. For if $Y \rightsquigarrow Z$, then $Y$ must precede $Z$ in the causal ordering. Thus, if $X_0 \rightsquigarrow X_1 \wedge \ldots \wedge X_{k-1} \rightsquigarrow X_k$, then $X_0$ must precede $X_k$ in the causal ordering, so $X_k$ cannot affect $X_0$. Thus, $\neg(X_k \rightsquigarrow X_0)$ holds. As we shall see, in a precise sense, C6 characterizes recursive models.

C6 can be viewed as a collection of axioms (actually, axiom schemes), one for each $k$. The case $k = 1$ already gives us $\neg(Y \rightsquigarrow Z) \vee \neg(Z \rightsquigarrow Y)$ for all variables $Y$ and $Z$. That is, it tells us that, for any pair of variables, at most one affects the other. However, just restricting C6 to the case of $k = 1$ does not suffice to characterize $\mathcal{T}_{\text{rec}}$, as the following example shows.

**Example 3.1:** Let $\mathcal{S} = (\emptyset, \{X_0, X_1, X_2\}, \{V_{X_0}, V_{X_1}, V_{X_2}\})$, where $V_{X_i} = \{0, 1, 2\}$ for $i = 0, 1, 2$, and let $T = (\mathcal{S}, \{F_{X_0}, F_{X_1}, F_{X_2}\})$, where $F_{X_i}$ is characterized by the equation

$$X_i = \begin{cases} 2 & \text{if } X_{i \ominus 1} = 1 \\ 0 & \text{otherwise} \end{cases}$$

and $\ominus$ is addition mod 3. It is easy to see that $T \in \mathcal{T}_{\text{uniq}}$: If any of the variables are set, the equations completely determine the values of all the other variables. On the other hand, if none of the variables are set, it is easy to see that $(0,0,0)$ is the only solution that satisfies all the equations. Moreover, in $T_{\vec{X} \leftarrow \vec{x}}$, the variable $X_i$ is 0 unless it is set to a value other than 0 or $X_{i \ominus 1}$ is set to 1. It easily follows that $X_i$ is affected only by $X_{i \ominus 1}$. A straightforward verification (or an appeal to Theorem 3.2 below) shows that $T$ satisfies all the axioms other than C6. C6 does not hold in $T$, since $T \models X_0 \rightsquigarrow X_1 \wedge X_1 \rightsquigarrow X_2 \wedge X_2 \rightsquigarrow X_0$. This also shows that $T$ is not recursive. However, the restricted version of C6 (where $k = 1$) does hold in $T$. A generalization of this example (with $k$ random variables rather than just 2) can be used to show that we cannot bound $k$ at all in C6; we need it to hold for all finite values of $k$. ∎

---

[2] In an earlier draft of this paper, where C1 and C2 were introduced, C1 was called "uniqueness". Galles and Pearl [1998] then adopted this name as well. In retrospect, this axiom really does not say anything about uniqueness. The axiom which does is D10, which will be discussed later.

[3] The assumption that $V_X$ and $\mathcal{X}$ are finite also necessary for the abbreviation $X \rightsquigarrow Y$ used in C6 to be in $\mathcal{L}_{\text{uniq}}$; however, as I show in the full paper, there is a variant of C6 that applies both for finite and infinite signatures.

Let $\text{AX}_{\text{uniq}}(\mathcal{S})$ consist of C0–C5 and MP; let $\text{AX}_{\text{rec}}(\mathcal{S})$ consist of C0–C4, C6, and MP. We could include C5 in $\text{AX}_{\text{rec}}(\mathcal{S})$; I did not do so because, as Galles and Pearl [1998] point out, it follows from C3 and C6. Note that the signature $\mathcal{S}$ is a parameter of the axiom system, just as it is for the language and the set of models. This is because, for example, the set $\mathcal{V}_X$ (which is part of $\mathcal{S}$) appears explicitly in C1 and C2.

**Theorem 3.2:** $\text{AX}_{\text{uniq}}(\mathcal{S})$ (resp., $\text{AX}_{\text{rec}}(\mathcal{S})$) is a sound and complete axiomatization for $\mathcal{L}_{\text{uniq}}(\mathcal{S})$ with respect to $\mathcal{T}_{\text{uniq}}(\mathcal{S})$ (resp., $\mathcal{T}_{\text{rec}}(\mathcal{S})$).

**Proof:** See the appendix. ∎

As I said in the introduction, Galles and Pearl [1998] prove a similar completeness result for causal models whose variables satisfy a fixed causal ordering. Given a total ordering $\prec$ on the variables in $\mathcal{X}$, consider the following axiom:

Ord. $Y_{\vec{x}w}(\vec{u}) = Y_{\vec{x}}(\vec{u})$ if $Y \prec W$

Since $\vec{x}$, $w$, and $\vec{u}$ are implicitly universally quantified in Ord, this axiom says that $\neg(W \rightsquigarrow Y)$ holds if $Y \prec W$. It follows that if $W \rightsquigarrow Y$, then $W \prec Y$. From this and the fact that $\prec$ is a total order, it is easy to see that Ord implies C6.

Galles and Pearl show that C1–C4 and Ord is a sound and complete axiomatization with respect to the class of structures satisfying Ord for their language ($\mathcal{L}_{\text{GP}}$). More precisely, Galles and Pearl take $A_C$ to consist of the axioms C1–C4 and Ord (but not C0 or MP), and show, in their notation, that $S \models \sigma$ implies $S \vdash_{A_C} \sigma$, where $S \cup \{\sigma\}$ is a set of formulas in $\mathcal{L}_{\text{GP}}$. There is an important subtle point worth stressing about their result: C1–C3, which are axioms in $A_C$, are not expressible in $\mathcal{L}_{\text{GP}}$ (since their statement involves disjunction and negation).

So what exactly is Galles and Pearl's result saying? They interpret $S \models \sigma$, as usual, as meaning that in all causal models satisfying $S$, $\sigma$ is true.[4] They interpret $S \vdash_{A_C} \sigma$ as meaning that $\sigma$ is provable from $S$ and the axioms in the axioms of $A_C$ "together with the rules of logic", which presumably means C0 and MP. It follows easily from Theorem 3.2 that their result is correct (see below), but is unlike typical soundness and completeness proofs, since the proof of $\sigma$ from $S$ will in general involve formulas in $\mathcal{L}_{\text{uniq}}$ that are not in $\mathcal{L}_{\text{GP}}$. (In particular, this will happen whenever C1–C3 are used in the proof.)

To see that Galles and Pearl's result follows from Theorem 3.2, define $S^*$ to be the formula in $\mathcal{L}_{\text{uniq}}(\mathcal{S})$ which is the conjunction of the formulas in $S$ (there can only be finitely many, since $\mathcal{L}_{\text{GP}}(\mathcal{S})$ itself has only finitely many distinct formulas), together with the conjunction of all the instances of the axiom Ord (again, there are only finitely many). Note that $S \models \sigma$ holds iff $\mathcal{T}_{\text{uniq}}(\mathcal{S}) \models S^* \Rightarrow \sigma$ (since the formulas in Ord guarantee that the only causal models that satisfy $S^*$ are recursive, and hence are in $\mathcal{T}_{\text{uniq}}(\mathcal{S})$). Thus, by Theorem 3.2, $S \models \sigma$ iff $\text{AX}_{\text{uniq}}(\mathcal{S}) \vdash S^* \Rightarrow \sigma$, and this latter statement is equivalent to $S \vdash_{A_C} \sigma$, as defined by Galles and Pearl. In fact, Theorem 3.2 shows that $\text{AX}_{\text{uniq}}(\mathcal{S}) + \text{Ord}$ gives a sound and complete axiomatization with respect to causal models satisfying Ord for the language $\mathcal{L}_{\text{uniq}}(\mathcal{S})$, which allows Boolean connectives. (Of course, Theorem 3.2 shows more, since it extends Galles and Pearl's result to $\mathcal{T}_{\text{rec}}(\mathcal{S})$ and $\mathcal{T}_{\text{uniq}}(\mathcal{S})$.) This suggests that $\mathcal{L}_{\text{uniq}}$ is a more appropriate language for reasoning about causality than $\mathcal{L}_{\text{GP}}$, at least for causal models in $\mathcal{T}_{\text{uniq}}$. $\mathcal{L}_{\text{GP}}$ cannot express a number of properties of causal reasoning of interest (for example, the ones captured by axioms C1–C3). When we use $\mathcal{L}_{\text{uniq}}$, not only is every formula in $\mathcal{L}_{\text{uniq}}$ valid in $\mathcal{T}_{\text{uniq}}$ provable from the axioms in $\text{AX}_{\text{uniq}}$, but the proof involves only formulas in $\mathcal{L}_{\text{uniq}}$.

What about $\mathcal{T}$? I have not been able to find a complete axiomatization for the language $\mathcal{L}_{\text{uniq}}$ with respect to $\mathcal{T}$. However, I do not think that finding a complete axiomatization for $\mathcal{L}_{\text{uniq}}$ with respect to $\mathcal{T}$ is of great interest, because $\mathcal{L}_{\text{uniq}}$ is simply not a language appropriate for reasoning about causality in $\mathcal{T}$. Because there is not necessarily a unique solution to the equations that arise in a causal model $T \in \mathcal{T}$, it is useful to be able to say both that there exists a solution with certain properties and that *all* solutions have certain properties. This is precisely what the language $\mathcal{L}^+$ lets us do.[5] As I now show, there is in fact an elegant sound and complete axiomatization for $\mathcal{L}^+$ with respect to $\mathcal{T}$.

Consider the following axioms:

---

[4] Although they do not say this explicitly, it is clear that they intend to further restrict to structures satisfying $S$ and Ord, for the fixed order $\prec$. Without this restriction, their result is not true.

[5] Note that $\mathcal{L}^+$ allows us to say that there is a unique solution for a random variable $X$ after setting some other variables. For example, $\langle \vec{Y} \leftarrow \vec{y} \rangle true \wedge [\vec{Y} \leftarrow \vec{y}](X(\vec{u}) = x)$ says that there are solutions to the equations when $\vec{Y}$ is set to $\vec{y}$ and, in all of them, $X$ is uniquely determined to be $x$ (if the exogenous variables are set to $\vec{u}$).

D0. All instances of propositional tautologies.

D1. $[\vec{Y} \leftarrow \vec{y}](X(\vec{u}) = x \Rightarrow X(\vec{u}) \neq x')$ if $x, x' \in V_X$, $x \neq x'$ (functionality)

D2. $[\vec{Y} \leftarrow \vec{y}] \vee_{x \in V_X} X(\vec{u}) = x$ (definiteness)

D3. $\langle \vec{X} \leftarrow \vec{x} \rangle (W(\vec{u}) = w \wedge \vec{Y} = \vec{y})$
$\Rightarrow \langle \vec{X} \leftarrow \vec{x}; W \leftarrow w \rangle (\vec{Y} = \vec{y})$ (composition)

D4. $[\vec{W} \leftarrow \vec{w}; X \leftarrow x](X(\vec{u}) = x)$ (effectiveness)

D5. $(\langle \vec{X} \leftarrow \vec{x}; Y \leftarrow y \rangle (W(\vec{u}) = w \wedge \vec{Z}(\vec{u}) = \vec{z}) \wedge$
$\langle \vec{X} \leftarrow \vec{x}; W \leftarrow w \rangle Y(\vec{u}) = y \wedge \vec{Z}(\vec{u}) = \vec{z})) \Rightarrow$
$\langle \vec{X} \leftarrow \vec{x} \rangle W(\vec{u}) = w \wedge Y(\vec{u}) = y \wedge \vec{Z}(\vec{u}) = \vec{z})),$
where $\vec{Z} = \mathcal{X} - (\vec{X} \cup \{W, Y\})$ (reversibility)

D6. $(X_0 \rightsquigarrow X_1 \wedge \ldots \wedge X_{k_1} \rightsquigarrow X_k) \Rightarrow \neg(X_k \rightsquigarrow X_0)$ (recursiveness)

D7. $([\vec{X} \leftarrow \vec{x}]\varphi \wedge [\vec{X} \leftarrow \vec{x}](\varphi \Rightarrow \psi)) \Rightarrow [\vec{X} \leftarrow \vec{x}]\psi$ (distribution)

D8. $[\vec{X} \leftarrow \vec{x}]\varphi$ if $\varphi$ is a propositional tautology (generalization)

D9. $\langle \vec{Y} \leftarrow \vec{y} \rangle true \wedge \vee_{x \in V_X}([\vec{Y} \leftarrow \vec{y}](X(\vec{u}) = x)$ if $Y = \mathcal{X} - \{X\}$ (unique solutions for $\mathcal{X} - \{X\}$)

D10. $\langle \vec{Y} \leftarrow \vec{y} \rangle true \wedge \vee_{x \in V_X}([\vec{Y} \leftarrow \vec{y}](X(\vec{u}) = x)$ (unique solutions)

D1–D6 are the analogues of C1–C6 in $\mathcal{L}^+$. D4 and D6 are just C4 and C6, with no changes at all. The other axioms are not quite the same though. For example, C1 is actually $[\vec{Y} \leftarrow \vec{y}](X(\vec{u}) = x) \Rightarrow \neg[\vec{Y} \leftarrow \vec{y}](X(\vec{u}) = x'))$ if $x \neq x'$. By Lemma 2.2, this is equivalent to D1 in $\mathcal{T}_{\text{uniq}}$; however, the two formulas are not equivalent in general. Similarly, C2 is $\vee_{x \in V_X}[\vec{Y} \leftarrow \vec{y}](X(\vec{u}) = x)$, which is closer to D10 than D2 (since the disjunction is outside the scope of the $[\vec{Y} \leftarrow \vec{y}]$. Again, D10 and D2 are equivalent in $\mathcal{T}_{\text{uniq}}$ (both are equivalent to C2 in this case) but, in general, D10 is stronger than D2. Only D2 and D9, both weaker than D10, hold in $\mathcal{T}$. The exact analogue of C3 would use $[]$ instead of $\langle \rangle$ and say $Y(\vec{u}) = y$ instead of $\vec{Y}(\vec{u}) = \vec{y}$. For completeness, it is necessary to have a vector of variables here. Using $[]$ instead of $\langle \rangle$ also results in a valid formula (and would not require a vector $\vec{Y}$). While the two variants are equivalent in $\mathcal{T}_{\text{uniq}}$, they are different in general, and the one given here is the more useful. (More precisely, with it we get completeness, while the version with $[]$ does not suffice for completeness.) Similarly, in D5, we use $\langle \rangle$ instead of $[]$, and add the extra clause $\vec{Z}(\vec{u}) = \vec{z}$. Both turn out to be necessary for soundness. In some sense, we can think of D1–D6 as capturing the "true content" of C1–C6, once we drop the assumption that the structural equations have a unique solution. D7 and D8 are standard properties of modal operators. D10 is what we need to capture the fact that the structural equations have unique solutions.

Let $AX^+$ consist of D0–D5, D7–D9, and MP (modus ponens); let $AX_{\text{uniq}}^+$ be the result of adding D10 to $AX^+$; let $AX_{\text{rec}}^+$ be the result of adding D6 to $AX_{\text{uniq}}^+$.

**Theorem 3.3:** $AX^+(\mathcal{S})$ (resp., $AX_{\text{uniq}}^+(\mathcal{S})$, $AX_{\text{rec}}^+(\mathcal{S})$) is a sound and complete axiomatization for $\mathcal{L}^+(\mathcal{S})$ with respect to $\mathcal{T}(\mathcal{S})$ (resp., $\mathcal{T}_{\text{uniq}}(\mathcal{S})$, $\mathcal{T}_{\text{rec}}(\mathcal{S})$).

**Proof:** See the appendix. ∎

## 4 DECISION PROCEDURES

In this section I consider the complexity of deciding if a formula is satisfiable (or valid). This, of course, depends on the language ($\mathcal{L}^+$, $\mathcal{L}_{\text{uniq}}$, or $\mathcal{L}_{\text{GP}}$) and the class of models ($\mathcal{T}_{\text{rec}}$, $\mathcal{T}_{\text{uniq}}$, $\mathcal{T}$) we consider. It also depends on how we formulate the problem.

One version of the problem is to consider a fixed signature $\mathcal{S} = (\mathcal{U}, \mathcal{X}, \{V_Y : Y \in \mathcal{U} \cup \mathcal{X}\})$, and ask how hard it is to decide if a formula $\varphi \in \mathcal{L}^+(\mathcal{S})$ (resp., $\mathcal{L}_{\text{uniq}}(\mathcal{S})$, $\mathcal{L}_{\text{GP}}(\mathcal{S})$) is satisfiable in $\mathcal{T}_{\text{rec}}(\mathcal{S})$ (resp., $\mathcal{T}_{\text{uniq}}(\mathcal{S})$, $\mathcal{T}(\mathcal{S})$). If $\mathcal{S}$ is finite (that is, if $\mathcal{X}$ and $\mathcal{U}$ are finite and $V_Y$ is finite for each $Y \in \mathcal{U} \in \mathcal{X}$), this turns out to be quite easy, for trivial reasons.

**Theorem 4.1:** If $\mathcal{S}$ is a fixed finite signature, the problem of deciding if a formula $\varphi \in \mathcal{L}^+(\mathcal{S})$ (resp., $\mathcal{L}_{\text{uniq}}(\mathcal{S})$, $\mathcal{L}_{\text{GP}}(\mathcal{S})$) is satisfiable in $\mathcal{T}_{\text{rec}}(\mathcal{S})$ (resp., $\mathcal{T}_{\text{uniq}}(\mathcal{S})$, $\mathcal{T}(\mathcal{S})$) can be solved in time linear in $|\varphi|$ (the length of $\varphi$ viewed as a string of symbols).

**Proof:** If $\mathcal{S}$ is finite, there are only finitely many causal structures in $\mathcal{S}$, independent of $\varphi$. Given $\varphi$, we can explicitly check if $\varphi$ is satisfied in any (or all) of them. This can be done in time linear in $|\varphi|$. Since $\mathcal{S}$ is not a parameter to the problem, the huge number of possible structures that we have to check only affects the constant. ∎

We can do even better than Theorem 4.1 suggests if $\mathcal{S}$ is a fixed finite signature. Suppose that $\mathcal{X}$ consists of 100 variables and $\varphi$ mentions only 3 of them. A causal model must specify the equations for all 100 variables. Is it really necessary to consider what happens to the 97 variables not mentioned in $\varphi$ to decide if $\varphi$ is satisfiable or valid? As the following result shows, if we restrict to models in $\mathcal{T}_{\text{uniq}}$, then we need to check only the variables that appear in $\mathcal{S}$; for models in $\mathcal{T}$, we need to allow one more variable.

More precisely, given a formula $\varphi$ and signature $\mathcal{S}$, let the signature $\mathcal{S}_\varphi = (\{U^*\}, \mathcal{X}', \{V_Y' : Y \in \mathcal{X}' \cup \{U^*\}\})$, where $\mathcal{X}'$ consists of the variables in $\mathcal{X}$ that appear in $\varphi$, $U^*$ is a fresh exogenous variable, not mentioned in $\mathcal{X}$ or $\mathcal{U}$, $V_X' = V_X$ for $X \in \mathcal{X}'$, and $V_{U^*}$ consists of all those tuples in $\times_{U \in \mathcal{U}} V_U$ that are mentioned in $\varphi$. Let $\mathcal{S}_\varphi^+$ be the same as $\mathcal{S}_\varphi$ except that it has one fresh endogenous variable $X^*$, with $V_{X^*} = \cup_{X \in \mathcal{X}'} V_X$.

**Theorem 4.2:** *A formula $\varphi \in \mathcal{L}^+(\mathcal{S})$ is satisfiable in $\mathcal{T}_{\text{rec}}(\mathcal{S})$ (resp., $\mathcal{T}_{\text{uniq}}(\mathcal{S})$) iff it is satisfiable in $\mathcal{T}_{\text{rec}}(\mathcal{S}_\varphi)$ (resp., $\mathcal{T}_{\text{uniq}}(\mathcal{S}_\varphi)$); $\varphi$ is satisfiable in $\mathcal{T}(\mathcal{S})$ iff it is satisfiable in $\mathcal{T}(\mathcal{S}_\varphi^+)$.*

**Proof:** See the full paper. ∎

Since Theorem 4.2 applies to all formulas in $\mathcal{L}^+(\mathcal{S})$, it applies *a fortiori* to formulas in $\mathcal{L}_{\text{uniq}}(\mathcal{S})$ and $\mathcal{L}_{\text{GP}}(\mathcal{S})$. Although stated only in terms of satisfiability, it is immediate that it also holds for validity. It tells us that without loss of generality, when considering satisfiability or validity, we need to consider only finitely many variables (essentially, only the ones that appear in $\varphi$, and perhaps one more).[6] In this sense, we can restrict to signatures with only finitely many variables without loss of generality. Note that this result does *not* tell us that we can restrict to finite sets of values for these variables without loss of generality.

Returning to the complexity of the decision problem, note that Theorem 4.1 is the analogue of the observation that for propositional logic, the satisfiability problem is in linear time if we restrict to a fixed set of primitive propositions. The proof that the satisfiability problem for propositional logic is NP-complete implicitly assumes that we have an unbounded number of primitive propositions at our disposal.

There are two ways to get an analogous result here. The first is to allow the signature $\mathcal{S}$ to be infinite and the second is to make the signature part of the input to the problem. The results in both cases are similar.

**Theorem 4.3:** *Given as input a pair $(\varphi, \mathcal{S})$, where $\varphi \in \mathcal{L}^+(\mathcal{S})$ (resp., $\mathcal{L}_{\text{uniq}}(\mathcal{S})$, $\mathcal{L}_{\text{GP}}(\mathcal{S})$) and $\mathcal{S}$ is a finite signature, the problem of deciding if $\varphi$ is satisfiable with respect to $\mathcal{T}$ (resp., (resp., $\mathcal{T}_{\text{rec}}$, $\mathcal{T}_{\text{uniq}}$) is NP-hard and in NEXPTIME (nondeterministic exponential time).*

**Proof:** It is quite straightforward to encode the satisfiability problem for propositional logic into the satisfiability problem for these logics. The details appear in the full paper. However, it is worth noting that a special encoding is needed for $\mathcal{L}_{\text{GP}}$, and the encoding makes crucial use of the fact that $V_X$ is finite for each endogenous variable $X$.

The upper bound is also straightforward: We simply guess a satisfying model and verify that it indeed satisfies $\varphi$. The reason we get nondeterministic *exponential* time rather than nondeterministic *polynomial* time is that the description of a causal model can be exponential in $|\varphi|$, and computing the solutions to the equations also takes exponential time. For example, if $\varphi$ mentions the variables $X_1, \ldots, X_{n+1}$, each of which has two possible values, then each of the equations $F_{X_1}, \ldots, F_{X_n}$ has $2^n$ possible inputs, and we have to say what the output is for each of them. ∎

---

[6] As I show in the full paper, the extra variable is necessary in the case of $\mathcal{T}(\mathcal{S})$.

One way of improving the upper bound of Theorem 4.3 would be to show that if a formula is satisfiable at all, it is satisfiable in a model that has a short (i.e., polynomial-length) description. In the case of $\mathcal{T}_{\text{rec}}$, this technique provides a matching upper bound.

**Theorem 4.4:** *Given as input a pair $(\varphi, \mathcal{S})$, where $\varphi \in \mathcal{L}^+(\mathcal{S})$ (resp., $\mathcal{L}_{\text{uniq}}(\mathcal{S})$, $\mathcal{L}_{\text{GP}}(\mathcal{S})$) and $\mathcal{S}$ is a finite signature, the problem of deciding if $\varphi$ is satisfiable with respect to $\mathcal{T}_{\text{rec}}(\mathcal{S})$ is NP-complete.*

**Proof:** See the full paper. ∎

I conjecture that it should be possible to similar NP-completeness results for $\mathcal{T}_{\text{uniq}}$ and $\mathcal{T}$, but I have not yet been able to do this.

Results similar to Theorems 4.3 and 4.4 hold for the case where $\mathcal{S}$ is an infinite signature. For example, as is shown in the full paper, slight modifications to the proof of Theorem 4.4 give us the following result.

**Theorem 4.5:** *If $\mathcal{S}$ is a fixed infinite signature, the problem of deciding if a formula $\varphi \in \mathcal{L}^+(\mathcal{S})$ (resp., $\mathcal{L}_{\text{uniq}}(\mathcal{S})$, $\mathcal{L}_{\text{GP}}(\mathcal{S})$) is satisfiable in $\mathcal{T}_{\text{rec}}(\mathcal{S})$ is NP-complete.*

One interesting difference is observed in the case of the language $\mathcal{L}_{\text{GP}}$. Having infinitely many possible values makes the decision problem *easier* in this case. In fact, even the implication problem for formulas in $\mathcal{L}_{\text{GP}}$ is in polynomial time.

**Theorem 4.6:** *If $\mathcal{S}$ is an infinite signature where for all but finitely many of the variables $X \in \mathcal{X}$, we have $|V_X| = \infty$, then we can decide if the formula $\varphi \in \mathcal{L}_{\text{GP}}(\mathcal{S})$ is satisfiable or valid with respect to $\mathcal{T}_{\text{rec}}(\mathcal{S})$ in time polynomial in $|\varphi|$. In fact, we can decide whether $\varphi_1 \Rightarrow \varphi_2$ is valid with respect to $\mathcal{T}_{\text{rec}}(\mathcal{S})$, for $\varphi_1, \varphi_2 \in \mathcal{L}_{\text{GP}}(\mathcal{S})$, in time polynomial in $|\varphi_1 \Rightarrow \varphi_2|$.*

**Proof:** See the full paper. ∎

## 5  CONCLUSION

I have provided complete axiomatizations and decision procedures for propositional languages for reasoning about causality. I have tried to stress the important role of the choice of language in both the axiomatizations and, more generally, in the reasoning process.

Both the models and the languages considered here are somewhat limited. For example, a more general approach to modeling causality would allow there to be more than one value of $X$ once we have set all the other variables. This would be appropriate if we model things at a somewhat coarser level of granularity, where the values of all the variables other than $X$ do not suffice to completely determine the value of $X$. I believe the results of this paper can be extended in a straightforward way to deal with this generalization, although I have not checked the details. For general causal reasoning, I believe we need a richer language, which includes some first-order features. I hope to return to the issue of finding appropriate richer languages for causal reasoning in future work.

## A  PROOFS

**Theorem 3.2:** *$AX_{\mathrm{uniq}}$ (resp., $AX_{\mathrm{rec}}$) is a sound and complete axiomatization for $\mathcal{L}_{\mathrm{uniq}}(\mathcal{S})$ with respect to $\mathcal{T}_{\mathrm{uniq}}(\mathcal{S})$ (resp., $\mathcal{T}_{\mathrm{rec}}(\mathcal{S})$).*

**Proof:** Soundness is proved by Galles and Pearl. To make the paper self-contained, I reprove the only non-obvious case—the validity of C5 in $\mathcal{T}_{\mathrm{uniq}}$.

Let $T \in \mathcal{T}_{\mathrm{uniq}}$ and suppose that $T \models Y_{\vec{x}w}(\vec{u}) = y \wedge W_{\vec{x}y}(\vec{u}) = w$. We want to show that $T \models Y_{\vec{x}}(\vec{u}) = y$. Since we are in $\mathcal{T}_{\mathrm{uniq}}$, there is a unique vector $\vec{v}_1$ that satisfies the equations in $T_{\vec{x}w}(\vec{u})$ and a unique vector $\vec{v}_2$ that satisfies the equations in $T_{\vec{x}y}(\vec{u})$. I claim that $\vec{v}_1 = \vec{v}_2$. By assumption, the $\vec{X}$, $Y$, and $W$ components of these vectors are the same ($\vec{x}$, $y$, and $w$, respectively). Now consider the theory $T_{\vec{x}yw}(\vec{u})$. I claim that $\vec{v}_1$ and $\vec{v}_2$ are both solutions to that theory. Note that for any variable $Z$ other than those in $\vec{X} \cup \{W, Y\}$, the equation $F_Z^{\vec{x}w,\vec{u}}$ for $Z$ in $T_{\vec{x}w}(\vec{u})$ is the same as the equations $F_Z^{\vec{x}y,\vec{u}}$ and $F_Z^{\vec{x}yw,\vec{u}}$ for $Z$ in $T_{\vec{x}y}(\vec{u})$ and $T_{\vec{x}yw}(\vec{u})$, respectively, except that in the first case, $w$ has been plugged in as the value of $W$, in the second case $y$ has been plugged in as the value of $Y$, and in the third case, both $w$ and $y$ have been plugged in. However, since $w$ and $y$ are the values of $W$ and $Y$, respectively, in both $\vec{v}_1$ and $\vec{v}_2$, and since these vectors satisfy both equation $F_Z^{\vec{x}w}$ and $F_Z^{\vec{x}y}$, they must also satisfy $F_Z^{\vec{x}wy}$. Since the equations for $T_{\vec{x}yw}(\vec{u})$ have a unique solution, we have that $\vec{v}_1 = \vec{v}_2$, as desired.

Next, I claim that $\vec{v}_1$ satisfies the equations in $T_{\vec{x}}(\vec{u})$. Again, as above, it is clear that it satisfies the equation for $Z \notin \vec{X} \cup \{W, Y\}$. A similar argument shows that it satisfies the equation for $Y$ in $T_{\vec{x}}(\vec{u})$, since $\vec{v}_1$ satisfies the equation for $Y$ in $T_{\vec{x}w}(\vec{u})$. Finally, a similar argument shows that it satisfies the equation for $W$ in $T_{\vec{x}}(\vec{u})$, since $\vec{v}_2 = \vec{v}_1$ satisfies the equation for $W$ in $T_{\vec{x}y}(\vec{u})$. Since the $Y$ component of $\vec{v}_1$ is $y$, it follows that $Y_{\vec{x}}(\vec{u}) = y$.

So much for soundness. For completeness, it suffices to prove that if a formula in $\mathcal{L}_{\mathrm{uniq}}$ is consistent with $AX_{\mathrm{uniq}}$ (resp., $AX_{\mathrm{rec}}$), then it is satisfied in a causal model in $\mathcal{T}_{\mathrm{uniq}}$ (resp., $\mathcal{T}_{\mathrm{rec}}$). I now give the argument in the case of $AX_{\mathrm{uniq}}$.

Suppose that a formula $\varphi \in \mathcal{L}_{\mathrm{uniq}}(\mathcal{S})$, with $\mathcal{S} = (\mathcal{U}, \mathcal{X}, \{V_Y : Y \in \mathcal{U} \cup \mathcal{X}\})$, is consistent with $AX_{\mathrm{uniq}}$. Consider a maximal consistent set $C$ of formulas that includes $\varphi$. (A maximal consistent set is a set of formulas whose conjunction is consistent such that any larger set of formulas would be inconsistent.) It follows easily from standard propositional reasoning (i.e., using C0 and MP only) that such a maximal consistent set exists. Moreover, from C1 and C2, it follows that for each random variable $X \in \mathcal{X}$ and vector $\vec{y}$ of values, there exists exactly one element $x \in V_X$ such that $X_{\vec{y}} = x \in C$. I now construct a causal model $T = (\mathcal{S}, F) \in \mathcal{T}_{\mathrm{uniq}}(\mathcal{S})$ that satisfies every formula in $C$ (and, in particular, satisfies $\varphi$).

A term $X_{\vec{Y} \leftarrow \vec{y}}(\vec{u})$ is *complete* (for $X$) if $\vec{Y}$ consists of all the variables in $\mathcal{X} - X$. Thus, $X_{\vec{Y} \leftarrow \vec{y}}(\vec{u})$ is a complete term if every random variable other than $X$ is determined. We use the complete terms to define the structural equations. For each variable in $X \in \mathcal{X}$, define $F_X(\vec{u}, \vec{y}) = x$ if $X_{\vec{y}}(\vec{u}) = x$, where $X_{\vec{y}}(\vec{u})$ is a complete term. This gives us a causal model $T$. Now we have to show that this model is in $\mathcal{T}_{\mathrm{uniq}}$ and that all the formulas in $C$ are satisfied by $T$.

I show that $X_{\vec{Y} \leftarrow \vec{y}}(\vec{u}) = x$ is in $C$ iff $T \models X_{\vec{Y} \leftarrow \vec{y}}(\vec{u}) = x$ by induction on $|\mathcal{X}| - |\vec{Y}|$. The case where $|\mathcal{X}| - |\vec{Y}| = 0$ follows immediately from C4, since then $X$ is in $\vec{Y}$. If $|\mathcal{X}| - |\vec{Y}| \neq 0$, we can assume without loss of generality that $X$ is not in $\vec{Y}$, for otherwise the result again follows from C4. Given this assumption, if $|\mathcal{X}| - |\vec{Y}| = 1$, the result follows by definition of the equations $F_X$.

For the general case, suppose that $|\mathcal{X}| - |\vec{Y}| = k > 1$. We want to show that there is a unique solution to the equations in $T_{\vec{Y} \leftarrow \vec{y}}(\vec{u})$ and that, in this solution, $X$ has value $x$. To see that there is a solution, we define a vector $\vec{v}$ and show that it is in fact a solution. If $W \in \vec{Y}$ and $W \leftarrow w$ is the assignment to $W$ in $\vec{Y} \leftarrow \vec{y}$, then we set the $W$ component of $\vec{v}$ to $w$. If $W$ is not in $\vec{Y}$, then set the $W$ component of $\vec{v}$ to the unique value $w^*$ such that $W_{\vec{Y} \leftarrow \vec{y}}(\vec{u}) = w^*$ is in $C$. (By C1 and C2 there is such a unique value $w$.) I claim that $\vec{v}$ is a solution to the equations in $T_{\vec{Y} \leftarrow \vec{y}}(\vec{u})$.

To see this, let $W$ be a variable in $\mathcal{X}$ not in $\vec{Y}$. Let $\vec{Y}' = \vec{Y}W$. By C3 and C4, for every variable $Z \in \mathcal{X} - \vec{Y}'$, we have $Z_{\vec{y}w^*}(\vec{u}) = z^*$. Since $|\mathcal{X}| - |\vec{Y}'| = k - 1$, by the inductive hypothesis, $\vec{v}$ is in fact the unique solution for $T_{\vec{y}w^*}(\vec{u})$. For every variable $Z$ in $\mathcal{X} - \vec{Y}'$, the equation $F_Z^{\vec{y}w^*,\vec{u}}$ for $Z$ in $T_{\vec{y}w^*}(\vec{u})$ is the same as the equation $F_Z^{\vec{y},\vec{u}}$ for $Z$ in $T_{\vec{y}}(\vec{u})$, except that $W$ is set to $w^*$. Thus, every equation in $T_{\vec{y}}(\vec{u})$ except possibly the equation $F_W^{\vec{y},\vec{u}}$ is satisfied by $\vec{v}$. To see that $F_W^{\vec{y},\vec{u}}$ is also satisfied by $\vec{v}$, simply repeat this argument above starting with another variable $W'$ in $\mathcal{X} - \vec{Y}$. (Such a variable must exist since $|\mathcal{X}| - |\vec{Y}|$ was assumed to be at least 2.)

It remains to show that $\vec{v}$ is the *unique* solution to the equations in $T_{\vec{y}}(\vec{u})$. Suppose there were another solution, say $\vec{v}'$, to the equations. Suppose that for each variable $W$ in $\mathcal{X} - \vec{Y}$, the $W$ component of $\vec{v}'$ is $w^{**}$. For some variable $Z$, we must have $z^{**} \neq z^{*}$. Since $Z_{\vec{y}}(\vec{u}) = z^{*}$, by assumption, it follows from C1 that $Z_{\vec{y}}(\vec{u}) \neq z^{**}$ is in $C$ (since $C$ is a maximal consistent set). It is also easy to see that for each $W$ in $\mathcal{X} - \vec{Y}$, the vector $\vec{v}'$ is also a solution to the equations in $T_{\vec{y}w^{**}}(\vec{u})$. Let $W$ be a variable other than $Z$ in $\mathcal{X} - \vec{Y}$. By the induction hypothesis, it follows that $W_{\vec{y}z^{**}}(\vec{u}) = w^{**}$ and $Z_{\vec{y}w^{**}}(\vec{u}) = z^{**}$ are both in $C$. By C5 (reversibility), $Z_{\vec{y}}(\vec{u}) = z^{**}$ is in $C$. But this contradicts the consistency of $C$.

This completes the proof in the case of $\mathcal{T}_{\text{uniq}}(\mathcal{S})$. Essentially the same proof works for $\mathcal{T}_{\text{rec}}$. We just need to observe that C6 guarantees that the theory we construct can be taken to be recursive. To see this, given a formula $\varphi$ consistent with $\mathcal{T}_{\text{rec}}$, consider a maximal set $C$ of formulas consistent with $\mathcal{T}_{\text{rec}}$ that contains $\varphi$. Let $T_C$ be the causal model determined by $C$, as above. The set $C$ also determines a relation $\prec$ on the exogenous variables: define $Y \prec Z$ if $Y \rightsquigarrow Z \in C$. It easily follows from C6 that the transitive closure $\prec^*$ of $\prec$ is a partial order: if $X \prec^* Y$ and $Y \prec^* X$, then $X = Y$. Any total order on the variables consistent $\prec^*$ gives an ordering for which $T_C$ is recursive. ∎

**Theorem 3.3:** $AX^+$ (resp., $AX^+_{\text{uniq}}$, $AX^+_{\text{rec}}$) is a sound and complete axiomatization for $\mathcal{L}^+(\mathcal{S})$ with respect to $\mathcal{T}(\mathcal{S})$ (resp., $\mathcal{T}_{\text{uniq}}(\mathcal{S})$, $\mathcal{T}_{\text{rec}}(\mathcal{S})$).

**Proof:** Soundness proceeds much as that of Theorem 3.2; I leave details to the reader. For completeness, we again proceed much as in the proof of Theorem 3.2. Because the proofs are so similar in spirit, I just sketch the proof for $AX^+$; the modifications for $AX^+_{\text{uniq}}$ and $AX^+_{\text{rec}}$ are left to the reader.

Again, given a formula $\varphi$ consist with $AX^+$, we consider a maximal consistent set of formulas containing $\varphi$ that is consistent with $AX^+$, and use it to construct a causal model $T$. Note that D9 suffices for this, because in defining $F_X(\vec{u}, \vec{y})$, we needed to know only the unique $x$ such that $[\vec{Y} \leftarrow \vec{y}](X(\vec{u}) = x)$ for $\vec{Y} = \mathcal{X} - X$, and D9 (together with D1) assures us that there is a unique such $x$. Again, we want to show that all the formulas in $C$ are satisfied by $T$.

Standard techniques of modal logic (using D0, D7, D8, and MP) can be used to show that $\langle \vec{Y} \leftarrow \vec{y}\rangle(\varphi_1 \vee \varphi_2) \in C$ iff either $\langle \vec{Y} \leftarrow \vec{y}\rangle\varphi_1 \in C$ or $\langle \vec{Y} \leftarrow \vec{y}\rangle\varphi_2 \in C$. From D2 it follows that $\langle \vec{Y} \leftarrow \vec{y}\rangle(\varphi \wedge X(\vec{u}) \neq x) \in C$ iff $\langle \vec{Y} \leftarrow \vec{y}\rangle(\varphi \wedge (\vee_{x' \in V_X - \{x\}} X(\vec{u}) = x') \in C$. From these two facts, it easily follows that it suffices to show that $\langle \vec{Y} \leftarrow \vec{y}\rangle(\vec{X}(\vec{u}) = \vec{x}) \in C$ iff $T \models \langle \vec{Y} \leftarrow \vec{y}\rangle(\vec{X}(\vec{u}) = \vec{x})$ for $\vec{X} = \mathcal{X} - \vec{Y}$. To do this, we proceed by induction on $|\mathcal{X}| - |\vec{Y}|$ again. The base case is dealt with using D4, as before. So assume that $k \geq 1$ and $|\mathcal{X}| - |\vec{Y}| = k + 1$. Suppose that $\langle \vec{Y} \leftarrow \vec{y}\rangle(\vec{X}(\vec{u}) = \vec{x}) \in C$. Let $X_1, X_2 \in \vec{X}$. Suppose that $X_1 \leftarrow x_1$ and $X_2 \leftarrow x_2$ are the assignments to $X_1$ and $X_2$ in $\vec{X} \leftarrow \vec{x}$. Let $\vec{X}' \leftarrow \vec{x}'$ and $\vec{X}'' \leftarrow \vec{x}''$ be the result of removing $X_1 \leftarrow x_1$ and $X_2 \leftarrow x_2$, respectively, from $\vec{X} \leftarrow \vec{x}$. By D3, both $\langle \vec{Y} \leftarrow \vec{y}; X_1 \leftarrow x_1\rangle(\vec{X}''(\vec{u}) = \vec{x}'')$ and $\langle \vec{Y} \leftarrow \vec{y}; X_2 \leftarrow x_2\rangle(\vec{X}'(\vec{u}) = \vec{x}')$ are in $C$. By the induction hypothesis, both of these formulas are true in $T$. By the soundness of D5, it follows that $T \models \langle \vec{Y} \leftarrow \vec{y}\rangle(\vec{X}(\vec{u}) = \vec{x}')$, as desired.

Conversely, suppose that $T \models \langle \vec{Y} \leftarrow \vec{y}\rangle(\vec{X}(\vec{u}) = \vec{x}')$. Then, since D3 is sound, we have that $T \models \langle \vec{Y} \leftarrow \vec{y}; X_1 \leftarrow x_1\rangle(\vec{X}''(\vec{u}) = \vec{x}'')$ and $T \models \langle \vec{Y} \leftarrow \vec{y}; X_2 \leftarrow x_2\rangle(\vec{X}'(\vec{u}) = \vec{x}')$. By the induction hypothesis, we have that both $\langle \vec{Y} \leftarrow \vec{y}; X_1 \leftarrow x_1\rangle(\vec{X}''(\vec{u}) = \vec{x}'')$ and $\langle \vec{Y} \leftarrow \vec{y}; X_2 \leftarrow x_2\rangle(\vec{X}'(\vec{u}) = \vec{x}')$ are in $C$. We now apply D5 to complete the proof. ∎

## Acknowledgments

I'd like to thank Judea Pearl for his comments on a previous version of this paper, as well as his generous help in providing pointers to the literature.